# Exploiting Context When Learning to Classify

Peter D. Turney

Knowledge Systems Laboratory, Institute for Information Technology
National Research Council Canada, Ottawa, Ontario, Canada, K1A 0R6
turney@ai.iit.nrc.ca

**Abstract.** This paper addresses the problem of classifying observations when features are context-sensitive, specifically when the testing set involves a context that is different from the training set. The paper begins with a precise definition of the problem, then general strategies are presented for enhancing the performance of classification algorithms on this type of problem. These strategies are tested on two domains. The first domain is the diagnosis of gas turbine engines. The problem is to diagnose a faulty engine in one context, such as warm weather, when the fault has previously been seen only in another context, such as cold weather. The second domain is speech recognition. The problem is to recognize words spoken by a new speaker, not represented in the training set. For both domains, exploiting context results in substantially more accurate classification.

## 1 Introduction

A large body of research in machine learning is concerned with algorithms for classifying observations, where the observations are described by vectors in a multidimensional space of features. It often happens that a feature is context-sensitive. For example, when diagnosing spinal diseases, the significance of a certain level of flexibility in the spine depends on the age of the patient. This paper addresses the classification of observations when the features are context-sensitive.

In empirical studies of classification algorithms, it is common to randomly divide a set of data into a testing set and a training set. In this paper, the testing set and the training set have been deliberately chosen so that the contextual features range over values in the training set that are different from the values in the testing set. This adds an extra level of difficulty to the classification problem.

Section 2 presents a precise definition of context. General strategies for exploiting contextual information are given in Section 3. The strategies are tested on two domains. Section 4 shows how contextual information can improve the diagnosis of faults in an aircraft gas turbine engine. The classification algorithms used on the engine data were instance-based learning (IBL) [1, 2, 3] and multivariate linear regression (MLR) [4]. Both algorithms benefit from contextual information. Section 5 shows how context can be used to improve speech recognition. The speech recognition data were classified using IBL and cascade-correlation [5]. Again, both algorithms benefit from exploiting context. The work presented here is compared with related work by other researchers in Section 6. Future work is discussed in Section 7. Finally, Section 8 presents the conclusion. For the two domains and three classification algorithms studied here, exploiting contextual information results in a significant increase in accuracy.

## 2 Definition of Context

This section presents a precise definition of context. Let $C$ be a finite set of classes. Let $F$ be an $n$-dimensional feature space. Let $\vec{x} = (x_0, x_1, ..., x_n)$ be a member of $C \times F$; that is, $(x_1, ..., x_n) \in F$ and $x_0 \in C$. We will use $\vec{x}$ to represent a variable and $\vec{a} = (a_0, a_1, ..., a_n)$ to represent a constant in $C \times F$. Let $p$ be a probability distribution defined on $C \times F$. In the definitions that follow, we will assume that $p$ is a discrete distri-bution. It is easy to extend these definitions for the continuous case.

**Primary Feature:** Feature $x_i$ (where $1 \leq i \leq n$) is a *primary feature* for predicting the class $x_0$ when there is a value $a_i$ of $x_i$ and there is a value $a_0$ of $x_0$ such that:

$$p(x_0 = a_0 | x_i = a_i) \neq p(x_0 = a_0) \tag{1}$$

In other words, the probability that $x_0 = a_0$, given $x_i = a_i$, is different from the probability that $x_0 = a_0$.

**Contextual Feature:** Feature $x_i$ (where $1 \leq i \leq n$) is a *contextual feature* for predicting the class $x_0$ when $x_i$ is *not* a primary feature for predicting the class $x_0$ and there is a value $\vec{a}$ of $\vec{x}$ such that:

$$p(x_0 = a_0 | x_1 = a_1, ..., x_n = a_n)$$
$$\neq p(x_0 = a_0 | x_1 = a_1, ..., x_{i-1} = a_{i-1}, x_{i+1} = a_{i+1}, ..., x_n = a_n) \tag{2}$$

In other words, if $x_i$ is a contextual feature, then we can make a better prediction when we know the value $a_i$ of $x_i$ than we can make when the value is unknown, assuming that we know the values of the other features, $x_1, ..., x_{i-1}, x_{i+1}, ..., x_n$.

The definitions above refer to the class $x_0$. In the following, we will assume that the class is fixed, so that we do not need to explicitly mention the class.

**Irrelevant Feature:** Feature $x_i$ (where $1 \leq i \leq n$) is an *irrelevant feature* when $x_i$ is *neither* a primary feature *nor* a contextual feature.

**Context-Sensitive Feature:** A primary feature $x_i$ is *context-sensitive* to a contextual feature $x_j$ when there are values $a_0$, $a_i$, and $a_j$, such that:

$$p(x_0 = a_0 | x_i = a_i, x_j = a_j) \neq p(x_0 = a_0 | x_i = a_i) \tag{3}$$

The primary concern here is strategies for handling context-sensitive features.

When $p$ is unknown, it is often possible to use background knowledge to distinguish primary, contextual, and irrelevant features. Examples of this use of background knowledge will be presented later in the paper.

## 3 Strategies for Exploiting Context

Katz *et al.* [6] list four strategies for using contextual information when classifying:

1. **Contextual normalization:** The contextual features can be used to normalize the context-sensitive primary features, prior to classification. The intent is to process context-sensitive features in a way that reduces their sensitivity to the context.
2. **Contextual expansion:** A feature space composed of primary features can be expanded with contextual features. The contextual features can be treated by the classifier in the same manner as the primary features.
3. **Contextual classifier selection:** Classification can proceed in two steps: First select a specialized classifier from a set of classifiers, based on the contextual features. Then apply the specialized classifier to the primary features.
4. **Contextual classification adjustment:** The two steps in strategy 3 can be reversed: First classify, using only the primary features. Then make an adjustment to the classification, based on the contextual features.

This paper examines strategies 1 and 2 (see Sections 4 and 5). A fifth strategy is also investigated:

5. **Contextual weighting:** The contextual features can be used to weight the primary features, prior to classification. The intent of weighting is to assign more importance to features that, in a given context, are more useful for classification.

The purpose of contextual normalization is to treat all features equally, by removing the affects of context and measurement scale. Contextual weighting has a different purpose: to prefer some features over other features, if they may improve accuracy.

## 4 Gas Turbine Engine Diagnosis

This section compares contextual normalization (strategy 1) with other popular forms of normalization. Strategies 2 to 5 are not examined in this section. The application is fault diagnosis of an aircraft gas turbine engine. The feature space consists of about 100 continuous primary features (engine performance parameters, such as thrust, fuel flow, and temperature) and 5 continuous contextual features (ambient weather conditions, such as external air temperature, barometric pressure, and humidity). The observations fall in eight classes: seven classes of deliberately implanted faults and a healthy class [7].

The amount of thrust produced by an engine is a primary feature for diagnosing faults in the engine. The exterior air temperature is a contextual feature, since the engine's performance is sensitive to the exterior air temperature. Exterior air temperature is not a primary feature, since knowing the exterior air temperature, *by itself*, does not help us to make a diagnosis. This background knowledge lets us distinguish primary and contextual features, without having to determine the probability distribution.

The data consist of 242 observations, divided into two sets of roughly the same size. One set of observations was collected during warmer weather and the second set was collected during cooler weather. One set was used as the training set and the other as the testing set, then the sets were swapped and the process was repeated. Thus the sample size for testing purposes is 242.

The data were analyzed using two classification algorithms, a form of instance-based learning (IBL) [1, 2, 3] and multivariate linear regression (MLR) [4]. IBL and MLR were also used to preprocess the data by contextual normalization [7].

The following methods for normalization were experimentally evaluated:
1. no normalization (use raw feature data)
2. normalization without context, using
   a. normalization by minimum and maximum value in the training set (the minimum is normalized to 0 and the maximum is normalized to 1)
   b. normalization by average and standard deviation in the training set (subtract the average and divide by the standard deviation)
   c. normalization by percentile in the training set (if 10% of the values of a feature are below a certain level, then that level is normalized to 0.1)
   d. normalization by average and standard deviation in a set of healthy baseline observations (chosen to span a range of ambient conditions)
3. contextual normalization (strategy 1), using
   a. IBL (trained with healthy baseline observations)
   b. MLR (trained with healthy baseline observations)

Contextual normalization was done as follows. Let $\vec{x}$ be a vector of primary features and let $\vec{c}$ be a vector of contextual features. Contextual normalization transforms $\vec{x}$ to a vector $\vec{v}$ of normalized features, using the context $\vec{c}$. We used the following formula for contextual normalization:

$$v_i = (x_i - \mu_i(\vec{c}))/\sigma_i(\vec{c}) \tag{4}$$

In (4), $\mu_i(\check{c})$ is the expected value of $x_i$ and $\sigma_i(\check{c})$ is the expected variation of $x_i$, as a function of the context $\check{c}$. The values of $\mu_i(\check{c})$ and $\sigma_i(\check{c})$ were estimated using IBL and MLR, trained with healthy observations (spanning a range of ambient conditions) [7].

Table 1 (derived from Table 5 in [7]) shows the results of this experiment.

Table 1: A comparison of various methods of normalization.

| Classifier | Normalization | no. correct | percent correct |
|---|---|---|---|
| IBL | none | 102 | 42 |
| IBL | min/max train | 101 | 42 |
| IBL | avg/dev train | 97 | 40 |
| IBL | percentile train | 92 | 38 |
| IBL | avg/dev baseline | 111 | 46 |
| IBL | IBL | 139 | 57 |
| IBL | MLR | 128 | 53 |
| MLR | none | 100 | 41 |
| MLR | min/max train | 100 | 41 |
| MLR | avg/dev train | 100 | 41 |
| MLR | percentile train | 74 | 31 |
| MLR | avg/dev baseline | 100 | 41 |
| MLR | IBL | 103 | 43 |
| MLR | MLR | 119 | 49 |

For IBL, the average score without contextual normalization is 42% and the average score with contextual normalization is 55%, an improvement of 13%. For MLR, the improvement is 7%. According to the Student *t*-test, contextual normalization is significantly better than all of the alternatives that were examined [7].

## 5 Speech Recognition

This section examines strategies 1, 2, and 5: contextual normalization, contextual expansion, and contextual weighting. The problem is to recognize a vowel spoken by an arbitrary speaker. There are ten continuous primary features (derived from spectral data) and two discrete contextual features (the speaker's identity and sex). The observations fall in eleven classes (eleven different vowels) [8].

For speech recognition, spectral data is a primary feature for recognizing a vowel. The sex of the speaker is a contextual feature, since we can achieve better recognition by exploiting the fact that a man's voice tends to sound different from a woman's voice. Sex is not a primary feature, since knowing the speaker's sex, *by itself*, does not help us to recognize a vowel. This background knowledge lets us distinguish primary and contextual features, without having to determine the probability distribution.

The data are divided into a training set and a testing set. Each of the eleven vowels is spoken six times by each speaker. The training set is from four male and four female speakers ($11 \times 6 \times 8 = 528$ observations). The testing set is from four new male and three new female speakers ($11 \times 6 \times 7 = 462$ observations). Using a wide variety of neural network algorithms, Robinson [9] achieved accuracies ranging from 33% to 55% correct on the testing set. The mean score was 49%, with a standard deviation of 6%.

Three of the five strategies in Section 3 were applied to the data:

Contextual normalization: Each feature was normalized by equation (4), where the context vector $\check{c}$ is simply the speaker's identity. The values of $\mu_i(\check{c})$ and $\sigma_i(\check{c})$ are estimated simply by taking the average and standard deviation of $x_i$ for the speaker $\hat{c}$. In a practical application, this will require storing speech samples from a new speaker in a buffer, until enough data are collected to calculate the average and standard deviation.

Contextual expansion: The sex of the speaker was treated as another feature. This strategy is not applicable to the speaker's identity, since the speakers in the testing set are distinct from the speakers in the training set.

Contextual weighting: The features were multiplied by weights, where the weight for a feature was the ratio of inter-class deviation to intra-class deviation. The inter-class deviation of a feature indicates the variation in a feature's value, across class boundaries. It is the average, for all speakers in the training set, of the standard deviation of the feature, across all classes (all vowels), for a given speaker. The intra-class deviation of a feature indicates the variation in a feature's value, within a class boundary. It is the average, for all speakers in the training set and all classes, of the standard deviation of the feature, for a given speaker and a given class. The ratio of inter-class deviation to intra-class deviation is high when a feature varies greatly across class boundaries, but varies little within a class. A high weight (a high ratio) suggests that the feature will be useful for classification. This is a form of contextual weighting, because the weight is calculated on the basis of the speaker's identity, which is a contextual feature.

Table 2 shows the results of using different combinations of these three strategies with IBL. These results show that there is a form of synergy here, since the sum of the improvements of each strategy used separately is less than the improvement of the three strategies used together ( $(58 - 56) + (55 - 56) + (58 - 56) = 3\%$ vs. $66 - 56 = 10\%$ ).

Table 2: The three strategies applied to the vowel data.

| strategy 1: contextual normalization | strategy 2: contextual expansion | strategy 5: contextual weighting | no. correct | percent correct |
|---|---|---|---|---|
| No  | No  | No  | 258 | 56 |
| No  | No  | Yes | 269 | 58 |
| No  | Yes | No  | 253 | 55 |
| No  | Yes | Yes | 272 | 59 |
| Yes | No  | No  | 267 | 58 |
| Yes | No  | Yes | 295 | 64 |
| Yes | Yes | No  | 273 | 59 |
| Yes | Yes | Yes | 305 | 66 |

The three strategies were also tested with cascade-correlation [5]. Because of the time required for training the cascade-correlation algorithm, results were gathered for only two cases: With no preprocessing, cascade-correlation correctly classified 216 observations (47%). With preprocessing by all three strategies, cascade-correlation correctly classified 236 observations (51%). This shows that contextual information can be of benefit for both neural networks and nearest neighbor pattern recognition.

## 6 Related Work

The work described here is most closely related to [6]. However, [6] did not give a precise definition of the distinction between contextual features (their terminology: parameters or global features) and primary features (their terminology: features). They examined only contextual classifier selection, using neural networks to classify images, with context such as lighting. They found that contextual classifier selection resulted in increased accuracy and efficiency. They did not address the difficulties that arise when the context in the testing set is different from the context in the training set.

This work is also related to work in speech recognition on speaker normalization [8]. However, the work on speaker normalization tends to be specific to speech recognition. Here, the concern is with general-purpose strategies for exploiting context.

## 7 Future Work

Future work will extend the list of strategies, the list of domains that have been examined, and the list of classification algorithms that have been tested. It may also be possible and interesting to develop a general theory of strategies for exploiting context.

## 8 Conclusion

The general problem examined here is to accurately classify observations that have context-sensitive features. Examples are: the diagnosis of spinal problems, given that spinal tests are sensitive to the age of the patient; the diagnosis of gas turbine engine faults, given that engine performance is sensitive to ambient weather conditions; the recognition of speech, given that different speakers have different voices; the classification of images, given varying lighting conditions. There is clearly a need for general strategies for exploiting contextual information. The results presented here demonstrate that contextual information can be used to increase the accuracy of classifiers, particularly when the context in the testing set is different from the context in the training set.

## Acknowledgments

The gas turbine engine data and engine expertise were provided by the Engine Laboratory of the NRC, with funding from DND. The vowel data were obtained from the University of California data repository (ftp ics.uci.edu, directory /pub/machine-learning-databases) [10]. The cascade-correlation [5] software was obtained from Carnegie-Mellon University (ftp pt.cs.cmu.edu, directory /afs/cs/project/connect/code). The author wishes to thank Rob Wylie and Peter Clark of the NRC and two anonymous referees of ECML-93 for their helpful comments on this paper.